# Predictive Modeling For Real-Time Personalized Health Monitoring in Muscular Dystrophy Management

Mohammed Akkaoui*
School of Information Science & Engineering
Dalian Polytechnic University
Dalian, China
akkaoui@xy.dlpu.edu.cn

Zlata Serigina
The School of International Education
Fujian Medical University
Fuzhou, China
zlataakkaoui@mail.ru

Zheng Ren
Department of Software and Big Data Technology
Dalian Neusoft University of Information
Dalian, China
renzheng@neusoft.edu.cn

Fang Yuan
Associate Professor
School of Engineering Practice & Innovation-Entrepreneurship Education
Dalian Polytechnic University
Dalian, China
fangy@dlpu.edu.cn

*Abstract*—Muscular Dystrophy is a group of genetic disorders that progressively affect the strength and functioning of muscles, thereby affecting millions of people worldwide. The lifetime nature of MD requires continuous follow-up care due to its progressive nature. This conceptual paper proposes an Internet of Things-based system to support the management of MD through remote, multi-dimensional monitoring of patients in order to provide real-time health status updates. Traditional methods have failed to give actionable data in real time, hence denying healthcare providers the opportunity to make evidence-based decisions. Technology-driven approaches are urgently needed to provide deep insights into disease progression and patient health. It aims to enhance treatment strategies, enabling patients to better manage their condition and giving healthcare professionals more confidence in their management decisions.

I. INTRODUCTION

Muscular Dystrophy (MD) is a group of inherited disorders that cause progressive weakening of the muscles of the shoulder and pelvic girdles. Such progressive deterioration in muscle strength significantly affects mobility and day-to-day functioning in MD patients. Hence, patients are often required to consult with the healthcare provider on a regular basis to monitor their health condition, which proves to be physically and logistically challenging, especially for patients who do not have access to specialty care.

With the progression of technology, the healthcare industry is progressively incorporating the Internet of Things (IoT) to improve patient care outcomes. Specifically, wearable sensors have emerged as essential instruments for the continuous monitoring of health conditions. These devices facilitate the ongoing assessment of critical metrics including muscle activity, fatigue, and mobility, thereby allowing for timely interventions and customized care strategies for patients with muscular dystrophy (MD).

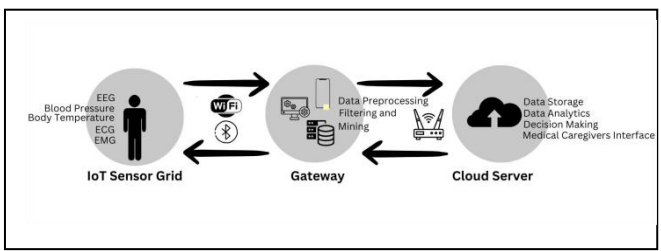

Fig 1: General IoT-based health tracking system

The proposed research shall present an IoT-based healthcare solution that will provide holistic monitoring and personalized care for patients with MD. This system will integrate wearable IoT devices with sensors for muscle activity, cardiac function, and environmental factors such as temperature and humidity. The collected data is then sent to a cloud-based platform where it is analyzed through advanced algorithms, and real-time updates of patient status are made available to healthcare professionals for timely interventions. The system also offers personalized recommendations about diet, exercise, and medication through a mobile app, thereby allowing patients to make informed decisions about their health.

However, such wearable devices cannot monitor particular biomarkers such as Creatine Kinase ;(CK) and transaminases (ALT/AST) that have an significant importance in assessing muscle and metabolic stress. Overcoming this drawback, the system uses point-of-care testing (POCT), the technique that is capable of efficiently and effectively evaluating small blood samples. CK has a major role in diagnosing muscle injury; its normal concentrations are usually between 20-200 U/L, while values over 1.000 U/L indicate significant damage. ALT and AST levels are also monitored using POCT to identify liver involvement or systemic stress, which is usual in patients with MD.

Sensors will be placed depending on the parameter to be measured for health. EMG sensors will be placed on large muscle groups like quadriceps, which are normally the first to show signs of atrophy. ECG sensors will monitor cardiac function, especially for arrhythmias that develop in advanced MD. Environmental sensors will track temperature and humidity, factors known to affect muscle fatigue. By monitoring these parameters, the system can detect abnormal trends, alerting healthcare providers and patients to enable early interventions.

It integrates the most advanced IoT framework for the management and monitoring of MD patients in real time. The comprehensive system will continue to capture data from different wearable sensors that track biomarkers, muscle function, and environmental factors. It will also use data aggregation, cloud-based analytics, and machine learning algorithms to detect anomalies and predict health risks.

The paper will discuss the design and implementation of the system, including sensor types and placement, and health metrics to be monitored. It will also look into the potential benefits for both patients and healthcare providers, especially in the areas of remote monitoring and personalized care plans. This research aims to show how IoT could revolutionize the management of chronic conditions such as MD and provide the impetus for future healthcare innovations.

## II. RELATED WORK AND MOTIVATION

Recent development in wearable sensor technologies has enabled remote monitoring of physical activity, gait, and other biometric data critical to patient health, especially for those suffering from neurological disorders. Giggins et al. (2017) undertook a comprehensive review of body-worn devices, but validity and reliability were covered for wearable sensors used in tracking physical activity in neurological patients. These findings are crucial to emphasize the significance of these devices in the determination of mobility and steps and the pertinent posture, further in developing interventions in a real-world setting. Consumer-grade sensors, such as accelerometers and gyroscopes, hold promise for continuous monitoring, but the reliability of data will depend on whether monitoring takes place in the clinical or daily life environment.

Probably the most applicable wearable sensors in fall prevention are those within elderly populations and people suffering from neurodegenerative diseases such as Parkinson's disease. Other fall prevention systems using wearable sensors include the monitoring of changes in physical activity and balance. Machine learning algorithms like SVM and CNN have been adapted to gait data in predicting falls, like the work done by Rajagopalan et al. (2021). Such systems have great potential in reducing fall-related risks and healthcare costs.

Moreover, as was evident from the systematic reviews, the wearable sensors applied to movement monitoring have expanded into other health areas, such as the monitoring of vital signs. For instance, wearable ECG, EDA, and PPG sensors track, on a continuous basis, the heart rate and skin temperature; this makes the monitoring of the patient so much holistic in both clinical and home settings. The use of multi-sensor systems for fall detection and in general health monitoring for people with neurological conditions becomes more widespread.

Despite these developments, challenges persist, particularly regarding sensor standardization, interoperability and interpretation of data accuracy in real-world deployment. Over these barriers lies the capability to achieve broader clinical impact and practical usability of such systems. I propose an integrated wearable sensor platform that merges the literature to date, with motion-tracking sensors, biometric data including heart rate and oxygen saturation, and non-invasive sensors estimating Creatine Phosphokinase (CPK) levels. The system shall improve early detection of fall risks and abnormal gait patterns through real-time alerts to caregivers or healthcare professionals. The system allows data aggregation, allowing a full strategy in the support of patient safety and timely interventions through cloud-based analytics for the remote management of patients. With this work, improvement in the existing systems will be realized by providing more accurate and reliable tools for continuously monitoring neurological patients at risk of falls, hence improving the health outcomes and quality of life in its users.

## III. SYSTEM ARCHITECTURE

The proposed MD patient monitoring and management system will be implemented using an advanced framework of IoT, which will provide its users with complete, real-time monitoring of all vital biomarkers, muscle function, environmental factors, and other important physiologic parameters. Wearable sensors will be used for collecting the data of essential health metrics continuously. Strong machine learning models will then analyze the collected data, which monitor disease progression and assess patient health toward making daily care management decisions.

### A. System Overview

The proposed system will attempt to serve as a comprehensive solution for all the varied needs of the MD patients. Muscular dystrophy is generally regarded as a group of inherited disorders that result in progressive muscle weakness and degeneration. Such conditions present problems in many arenas, particularly locomotion, respiratory health, and cardiac well-being. Our IoT-based solution embeds special sensors to monitor key parameters relevant to MD patients: muscle enzymes like CPK and ALT/AST, electromyographic signals, respiratory function, and cardiovascular health. Moreover, environmental factors such as temperature and humidity, which generally aggravate the condition, are monitored. The system not only enables the continuation of monitoring but also uses machine learning algorithms for predictive insights, which could help forecast the progression of a disease and guide day-to-day health management decisions (Figure 2).

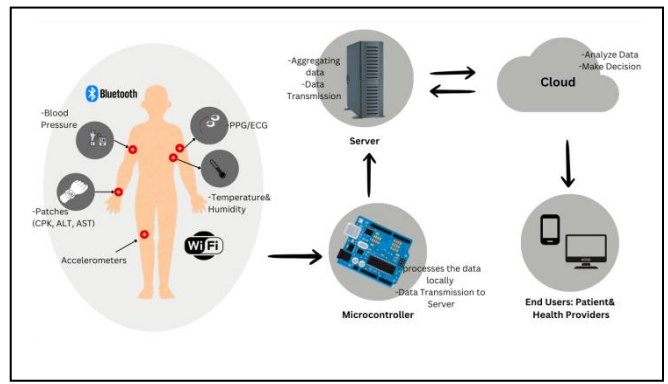

Fig 2: The IoT-based MD patients health monitoring system architecture

### B. Sensor Architecture

It incorporates multiple biometric and environmental sensors in a wearable sensor system-largely, placed across a patient's body to monitor disease-specific parameters.

Electromyography Sensors: These are placed over major muscle groups to measure the electrical activity within the muscles, crucial in early detection of muscle weakness and degeneration in MD patients.

CPK Monitoring: With wearable sensors still scarce for the measurement of CPK, point-of-care testing is going to be resorted to. Enabling CPK measurements that take only a few small blood samples—for example, from a finger prick—this method consequently brings results very quickly, ideal for frequent monitoring. This would thus enable health professionals to quickly determine muscle damage and make appropriate corresponding therapeutic interventions without the need for time-consuming laboratory infrastructure.

ALT/AST Monitoring: Similar to CPK, the levels of ALT and AST will be monitored using POCT devices. These allow quick measurement of liver enzymes in very small blood samples and provide healthcare providers with real-time data regarding metabolic stress or liver involvement. This assures timely intervention when levels of these enzymes exceed safe thresholds.

Cardiorespiratory Sensors: These include the PPG sensors, either wrist-mounted or chest-mounted, in order to provide information about vital signs such as heart rate, SpO2, and respiration rate. Those parameters are going to be relevant in assessing fatigue, respiration function, and cardiovascular function—all functions that deteriorate in MD patients.

Temperature and Humidity Sensors: Environmental sensors also track the room temperature, level of humidity, and air quality. Severe conditions may increase MD symptoms, therefore leading to increased fatigue and muscle stiffness. Movement sensors: Accelerometers and gyroscopes attached to the trunk or limbs measure balance, gait, and locomotion. Such may be a critical help in the prevention of falls and allow tracking of decline in mobility, often one of the features of MD that improves over time.

### C. Gateway Architecture and Communication

The system architecture will manage muscular dystrophy using IoE-based wearable sensors connecting to the central gateway integrated with cloud technology for data analytics and decision-making. Biomarker monitoring, environmental issues, and physical activity are all anchored within validated biomedical equations and clinical standards in order to make sure that all the data is valid and actionable.

Each of these sensors sends its reading to a local gateway device, which can be a smartphone or even an IoT-specific gateway. The gateway aggregates, filters, and performs preprocessing on the data before pushing it to the cloud. Preprocessing at the gateway, though it does reduce bandwidth consumption and latency by filtering out any unrequired data—like high-frequency EMG signals—is done using FFT, or Fast Fourier Transform, in order to detect abnormal muscle contractions. The gateway identifies certain thresholds, such as CPK levels above 1000 U/L or SpO2 falling below 90%, which require immediate attention. These critical events, so identified, are prioritized and uploaded to the cloud for further analysis.

**Data Transmission and Encryption:**

All the data communication between the iCare-1300 POCT device, wearable sensors, the gateway, and the cloud is encrypted with AES-256 for high security of the patient's information. With seamless handover between Wi-Fi and 4G/5G networks, this platform provides the continuum of monitoring even in the toughest urban and rural environments.

| Device Type | Data Type | Transmission Method | Frequency |
|---|---|---|---|
| iCare-1300 POCT Device | CPK, ALT, AST Results | Wi-Fi | Daily or immediately during abnormal muscle activity |
| Wearable Sensors (EMG, ECG, SpO2) | Muscle Activity, Cardiac Performance, Oxygen Levels | Bluetooth Low Energy (BLE) | Continuous; uploads to gateway |
| Environmental Sensors | Temperature, Humidity | Bluetooth Low Energy (BLE) | Every 30 minutes |
| Cloud Platform | Aggregated Sensor Data | WiFi/Cellular (4G/5G) | Continuous; real-time alerts for deviations |

Fig 3: This table summarizes the data transmission methods and frequencies used in the iCare-1300 POCT device and associated sensors.

This entire system promises secure data transfer from one device to another using redundant communication channels, making the design conclusively reliable. Integrating real-time biomarker monitoring and continuous environment and physiological parameter monitoring of patients will allow for prompt interventions, allowing patients to take a more proactive role in their health and well-being.

### D. Cloud-Based Analytics and Platform Integration

It collects and analyzes patient data from various sources against clinical benchmarks in the cloud. The system architecture supports an in-the-moment decision based on predictive analytics and historical trends. It allows healthcare providers to respond more fittingly to conditions in the present moment, predicting future health risks accurately according to past data.

It aggregates data from multiple patients, enabling broader trend analysis that will help refine treatment protocols and enable proactively intervention-oriented methods. Engineered for interoperability and scalability, this platform integrates well with EHRs and medical devices to orchestrate a unified healthcare ecosystem.

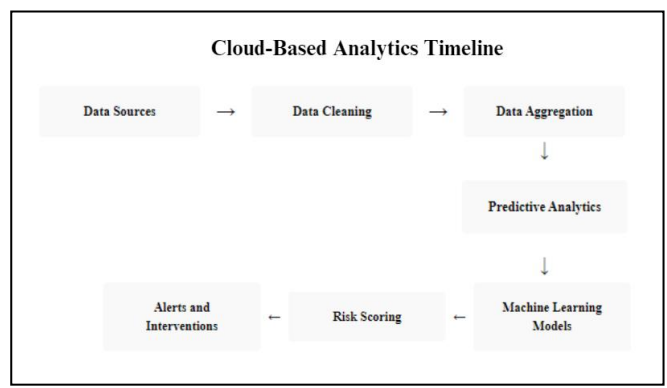

Fig 4: This timeline summarizes the components involved in cloud-based analytics and predictive modeling for muscular dystrophy management.

The system utilizes a range of powerful machine learning models, including Random Forests, Support Vector

Machines (SVMs), ARIMA, and LSTM networks, toward deeper analysis of patient data patterns, rendering real-time insights. Such models allow for the possibility of continued monitoring and proactive intervention through focusing on key health metrics crucial for the management of muscular dystrophy (MD).

Such signal processing techniques, such as Fast Fourier Transform (FFT), analyze data in EMG for detecting abnormal muscle contractions and frequency spikes that might indicate early signs of muscle degeneration. The machine learning models, once getting labeled data, help in mapping these abnormalities to MD progression so that clinicians can really detect muscle atrophy and it is clinically significant. This leads to early detection and, consequently, timely intervention with a change in the treatment plan accordingly. Similarly, time series models, which could be ARIMA or LSTM networks, analyze historical CPK data to make predictions of the future trends. Any models indicating that the CPK levels are to surpass 200 U/L, healthcare providers are immediately warned and can modify care either by changing medications or increasing physiotherapy to try to mitigate muscle damage. While ARIMA is ideal in following linear trends, LSTM networks capture non-linear patterns and sudden data spikes more accurately. Considering the unpredictable nature of muscular conditions, LSTM networks could be more flexible and accurate.

This integration of real-time analytics with predictive modeling enables the patient and healthcare provider to be informed and supportive in the proactive management of muscular conditions, enhancing quality of life for the patient. It creates personalized treatment plans based on the analytics about each patient's individual needs.

The RiskScore employs a new formula that has more weight towards CPK since it happens to be the most sensitive marker for muscle damage. This formulation ensures intervention happened before the overall risk score approaches critical levels. The present formula as follows:

(1)

$$RiskScore = 0.5 \times \frac{CPK}{1000} + 0.25 \times \frac{ALT}{100} + 0.25 \times \frac{AST}{100} + 0.1 \times EMG\ Amplitude\ (mV)$$

Creatine Phosphokinase is a very sensitive marker of muscle damage, especially in Muscular Dystrophy. The higher its value is, the more severe the injury to muscles, all the way to 1,000 U/L and above. It is heavily weighted at 0.5 in risk assessments. Other liver enzymes such as ALT and AST also rise with the concomitant muscle damage impinging on the overall systemic stress.

Their weights in the RiskScore are 0.25 each, as their contribution to overall health should not overshadow CPK, the representative of neuromuscular status. The amplitude of the EMG is a measure of strength below 0.5 mV functional decline and is weighed 0.1 in the RiskScore.

The final step uses heart rate and SpO2 values in predicting oxygen delivery to tissues and flags hypoxemic events at critical thresholds of SpO2 below 90% or heart rate above 120 bpm.

It also tracks various biomarkers and environmental factors through wearable sensors and marks early alerts in case of significant deviations. It offers timely interventions customized to the immediate needs of a patient.

Biochemical Markers A high CPK alert is if an elevation of more than 500 U/L in a 24-hour period or rises more than 200 U/L for an entire day require emergency treatment. If ALT and AST have values over 140 U/L and 100 U/L respectively this shows an invasive systemic or metabolic disease process and should be consulted with a physician. EMG amplitude value below 0.5 mV shows muscle atrophy and an early change in the physiotherapy settings must be made.

Cardiorespiratory metrics include heart rate variability below 20 ms, indicating autonomic dysfunction and the need for rest or relaxation. SpO2 levels below 90% indicate respiratory distress, requiring oxygen supplementation or reduced activity.

More importantly, high temperature over 30°C and high humidity over 70% enhance the muscle fatigue, so the system starts alerting for less strenuous activities with increased hydration. For exercise, it triggers resting alert to prevent muscle tear as soon as it captures sustained higher movement through accelerometers. One such feature fall detection alerts the caregivers on time which may in turn minimize further damage, depending upon acceleration or orientation changes.

This system provides real-time alerts based on evidence-based guidelines from AANEM, ACR, ESC, ATS, and CDC to give timely and evidence-based interventions for optimized patient health.

**Thresholds and Threshold Points for Decision Making:**

RiskScore > 3/10: Moderate Risk A moderate risk level threshold is set. Notifications at this stage went out to both the caregiver and healthcare provider, whose monitoring frequency increased to daily check-ins in order to catch upward trends early. The clinical guideline cited in the management of MD stated that the early detection of worsening biomarker trends preempts rapid deterioration and optimizes treatment.

RiskScore > 6/10: High Risk Scores of 6/10 or above put the individual in a high-risk category that must be intervened immediately. At this point, the scale at which CPK, ALT/AST, and/or EMG amplitude is rising associated with the above parameters confers radical transition in disease of muscle and systemic health. Such approaches are introduced according to research on MD, which points to an increase in biomarkers with fast progression and a higher tendency to increase in hospitalization. These weights and risk thresholds are calibrated in line with a clinically aligned approach to monitoring MD, bringing together biochemical markers in combination with functional assessment for effective and timely interventions.

## IV. System Design and Implementation: Cloud-Based Predictive Modeling and Analytics

Predictive analytics in healthcare is a nascent discipline that makes use of data science in making health outcome forecasts, identifying risk factors, and informing clinical decisions. The great benefit of predictive modeling offers a proactive approach to the care and management of chronic diseases, where monitoring is normally confined to periodic visits and post-hoc analyses through the constant processing of up-to-date patient information. Such a paradigm shift is going to be of growing importance when managing complex and progressive conditions such as Muscular Dystrophy.

Muscular Dystrophy is a genetic disorder wherein the person suffers from progressive muscle degeneration and weakness, affecting locomotion, respiration, and the quality of life of a patient. Such a gradually progressing, with variable severity in different patients' disease, presents unique challenges to healthcare modalities. Traditional MD patient monitoring was done through periodic clinical visits where the health care providers would assess the symptoms and disease advancement with regards to snapshots of health. This modality cannot catch the subtlety in disease evolution when critical changes fall between appointments.

Predictive modeling fills that gap by giving a seamless, data-driven view of patient health. Such models run on the data collected by wearable IoT devices that track biomarkers, such as Creatine Phosphokinase Levels, electromyographic signals, heart rate, environmental humidity, and temperature in order to provide real-time insight into the condition of the patient. Such models process vast numbers of data in order to detect subtle patterns and predict the risk-fraught that would otherwise go unnoticed both by healthcare providers and patients. Analyzing such patterns shall, therefore, allow predictive modeling to forecast critical health events—say, sudden muscle deterioration or respiratory complications—and thus provide timely alerts for the implementation of preventive measures.

In a nutshell, predictive modeling is the structured methodology that converts raw data into actionable insight. Firstly, a collection of raw data from various sources is done, including continuous biometrics from wearable sensors, historical medical records, and relevant environmental data. This way, data integration creates a complete picture of the patient's health condition, allowing for a better and more precise understanding of the factors involved that alter the course of disease prediction. Thus, preprocessing is done: cleaning, normalization, and extraction of meaningful features from raw data. Preprocessing cleans the raw data so that it can be fed into the prediction models, which are reliable, noise-free, and thus more accurate and robust in analysis. Model selection is a critical step; indeed, there is a large variety of choices of algorithms, each tuned for MD management needs, including time-series forecasting models such as ARIMA and LSTM and classification models including Support Vector Machines and Random Forests. Time-series models are very good at following up changes over time within biomarkers, including CPK levels, while classification models may classify patients according to risk level in order to drive appropriate interventions.

Each model learns from patterns in the historical data on MD progression and is validated based on sound techniques that assure its reliability in real-world settings.

These models would then be deployed on the cloud from where they will continue monitoring real-time data generated by patients' wearables. During this phase of predictive analytics, the streaming data is matched with predetermined thresholds and historical trends in real time.

Such dynamic, real-time analysis can identify variances that show that an adverse event may be occurring as a result of a decline in muscle function, respiratory distress, or other complications. For example, a predictive model might show a gradual elevation in CPK levels over several days, warning of an impending muscle breakdown. Early warning gives the wink and nod to the healthcare providers to take interventions sooner, probably adjusting medications or recommending physical therapy in order to reduce further damage. The system also relies on a mechanism of RiskScore beside the risk prediction. It performs dynamic evaluation based on various measures of

The health condition of the patient. To a large extent, this risk-scoring system concentrates on biomarkers like CPK and ALT/AST levels for muscle damage and systemic stress, respectively, in consideration of other parameters like EMG amplitudes, $SpO_2$ level, heart rate, and environmental conditions.

The RiskScore allows for an overall current risk of the patient to be created, especially one that automatically sets alerts to both health care providers and patients when it has already reached the threshold for moderate or high risk. This will allow timely interventions on the part of the health care providers before the condition of a patient deteriorates further and the patients can make an informed decision concerning their care.

The other salient feature of the predictive modeling framework is a feedback loop in continuing to learn about the model. As each day passes, more data accrues, and actual outcomes are observed; the model continues to refine its algorithms until it becomes absolutely—exceedingly—good at pattern recognition and predicting health events with increasing precision. The system is adaptive, which means that as the disease progresses or environmental factors change, the system remains sensitive to the changed health needs of the patient. That mechanism of feedback allows the model to not only enhance its accuracy of prediction over time but also personalize it for each individual, hence a truly patient-centric solution.

The following sections walk through predictive modeling in a sequential manner, including the main steps: data collection and integration, data preprocessing, model selection and training, real-time prediction and risk scoring, feedback, and optimization of the system. Each of the following subsections addresses with some degree of detail exactly how these components put together build a robust, adaptive, and personable health monitoring system. Figure 5: General Predictive Modeling Framework: It is an example of what has been outlined in much greater detail, from data acquisition to real-time monitoring and feedback. .

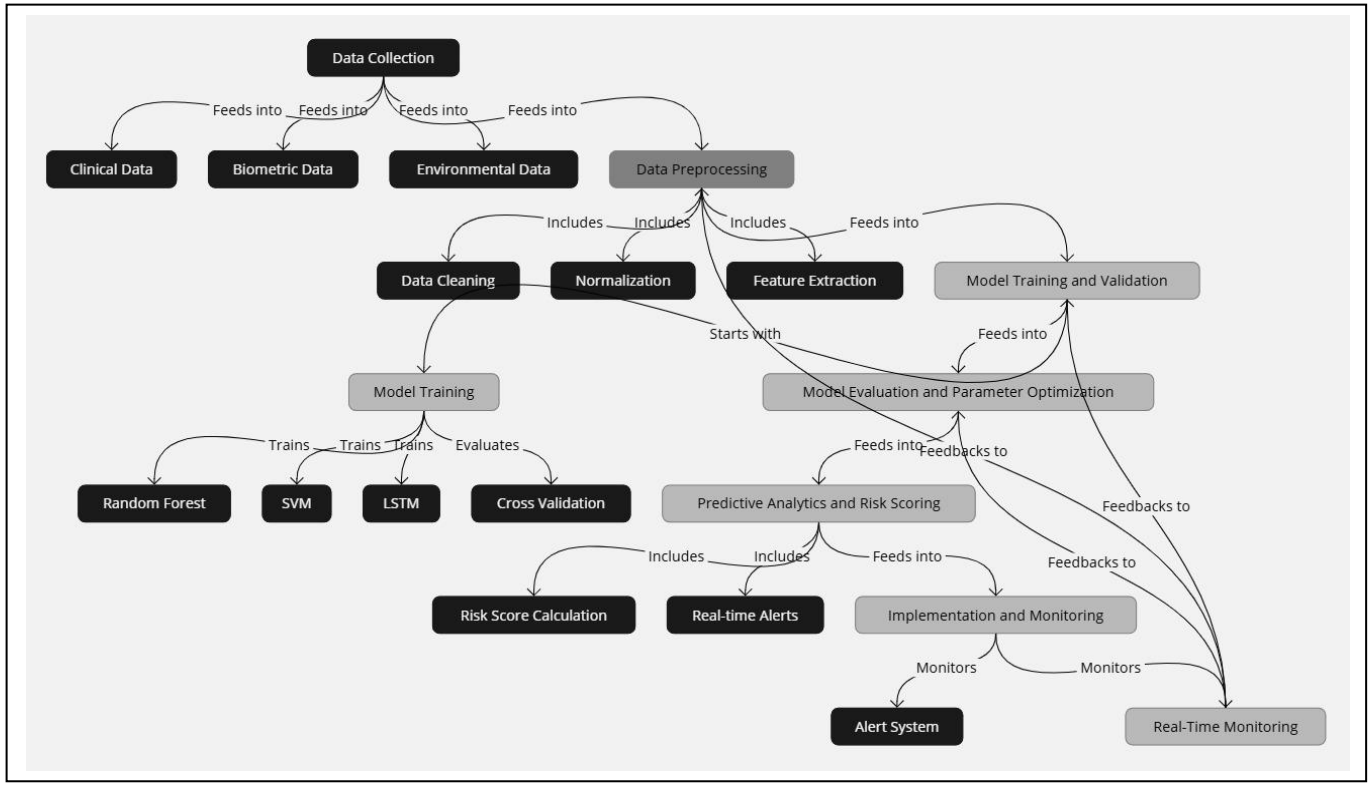

Fig 5: Overview of the predictive modeling framework

Predictive modeling of Muscular Dystrophy management really comes alive with a holistic real-time picture of a patient's health through the integration of various and different sources of data. The model will work on a systematic collection, integration, and storage of data of several types, from clinical information to biometric signals and environmental factors. This section shows the kind of data that is collected, methods of integration, and challenges associated therewith.

These data from the different sources need to be stored in a centralized system for analysis. It calls for highly skilled cloud infrastructure that can handle large volumes of data coming from multiple sensors and databases in a manner that is secure. Cloud storage means all the data collected is easily accessible and real-time for easy analysis, be it from the clinical records, wearable sensors, or environmental devices.

Synchronization protocols ensure that data does not experience fragmentation. All data are integrated from the wearable sensors, POCT devices, and clinical databases into one dataset. This form of integration ensures that all the predictive models shall have up-to-date and holistic information. Therefore, in real-time analytics, this synchronization is very important, where data should not be delayed coming from different sensors and devices.

Therefore, such data, when collected from disparate sources, naturally contains inconsistencies, missing values, and noises that generally affect the performance of predictive models. In general, data cleaning includes filtering out irrelevant information, treating missing values either by imputation or exclusion, and correcting errors in the input data so high-quality input can be performed for analysis.

From the raw data collected, relevant features must then be extracted in order to make a capture of the most important health indicators that will go to predictive modeling. In the example of feature extraction for EMG data, patterns of frequency and amplitude that reflect early-stage degeneration could be considered.

**Model Selection for Predictive Modeling in MD Management**

In this paper, several machine-learning models are under consideration for the prediction of disease progression and health risks related to the MD patients. Model choice was based on a need for high-accuracy real-time predictions, together with complexity in MD's progression and handling multiple patient data types: biometric, clinical, and environmental. Because of that, the major reasons for the selection of the applied models in the current research are the following:

*1. Complexity of Disease Progression*

The MD is a progressive and heterogeneous disease; the course of evolution may be very diverse among patients. Of course, this variability would require a very flexible yet segmented estimation model in pinpointing minute changes in disease markers over time. Hence, time-series models such as the ARIMA and LSTM network are best for this purpose.

Most of these models do indeed capture temporal dependencies that are, at their core, of utmost importance to forecast future health outcomes from past and current data concerning patients. The ARIMA models are really very efficient in capturing linear trends and seasonal patterns within time-series data. Because MD progression follows sometimes somewhat predictable patterns—for example, gradual rises in CPK levels, general muscle weakness—ARIMA's ability to predict future values from a given past is of value. That offers forecasted levels of CPK or other biomarkers for proactive healthcare intervention..

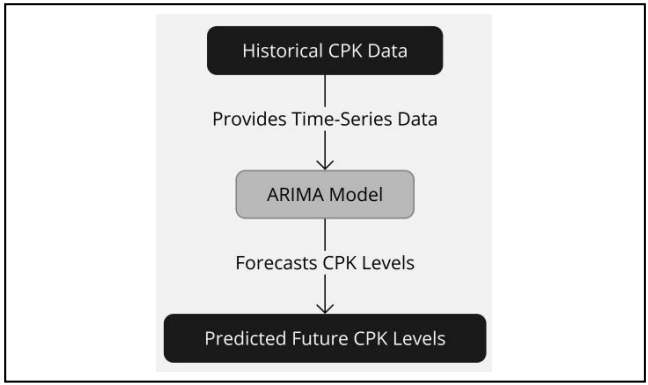

Fig 6: ARIMA Model for Predicting CPK Levels Over Time in MD Patients

Being a special kind of RNN, the LSTM networks have more appropriateness in modeling nonlinear patterns and long-range dependencies characteristic of time-series data. In most instances, there exists a very complex interaction between these numerous agents—medication, muscular damage, and environmental conditioning—that may induce a nonlinear effect on MD progression. The resulting models, therefore, have the ability to understand relationships between propositions and make a precise prediction of the future course of a particular disease, especially when sudden changes or irregularities have appeared in health indicators.

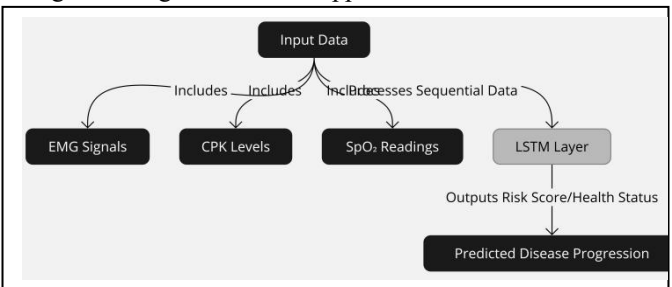

Fig 7: LSTM Model for Predicting MD Disease Progression

### 2. Data Complexity and Multivariate Nature

Such data variables for collecting information related to the management of MD include biometric signals like EMG and $SpO_2$, biochemical markers representing CPK levels and ALT/AST, and environmental data including humidity and temperature. Dealing with such a very broad range of such data variables being handled all at once, interfacing among them is a very challenging task.

In fact, models such as Random Forests and SVM are very helpful to them in the light of the following:

Random Forest is a significant method in ensemble learning that builds many decision trees and combines their outputs. The algorithm has reasonable robustness to deal with high-dimensional data with many features and is less sensitive to over-fitting compared with other models, for example, single decision trees. Here, the model can analyze the relationships between different features like $SpO_2$ levels, heart rate, temperature, and other features, and it also can estimate the contribution of the features to the health risk of patients.

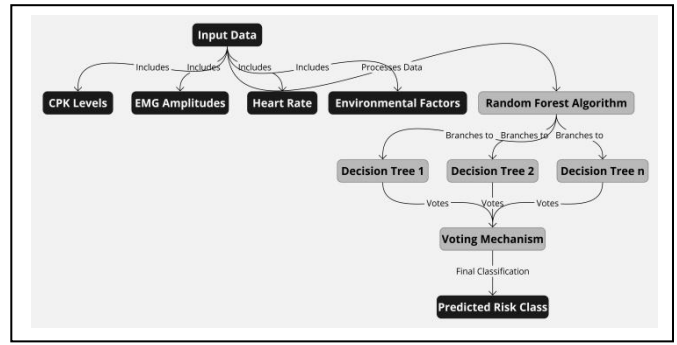

Fig 8: Random Forest for Risk Classification in MD Patients

Random Forest then outputs feature importance scores, allowing for the identification of the biomarkers or environmental factors that most affect MD progression; therefore, these should be the highest priorities for health professionals in early intervention.

SVM generally performs well in classification in many cases when dealing with highly nonlinear and not linearly separable datasets. The management of motion disorders using SVM to classify patients at any particular risk category, low, medium, or high, for disease progression based on synthesis of biomarkers, movement data, and environmental factors, is appropriate in most cases. This ability is crucial when dealing with complex multidimensional datasets, where the relationship among the different features may not be readily understood.

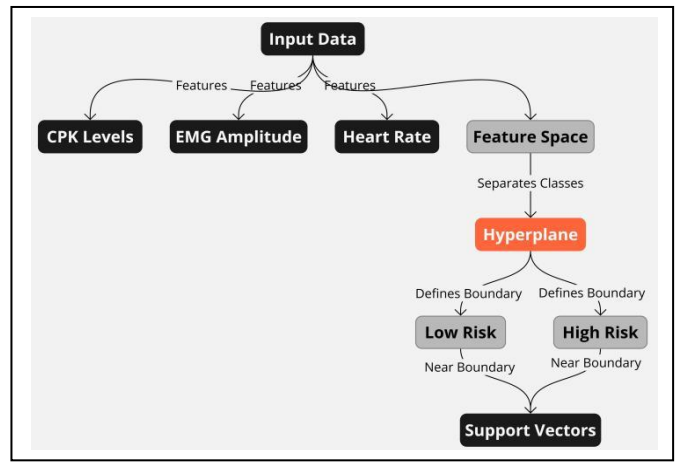

Fig 9: SVM for Risk Classification in MD Patients

In the health care setting, not only correct predictions are important but also actionable insights would be presented to the health care provider. Techniques such as Random Forests and SVMs are highly interpretable in the sense that the driving factors will be known to the health professional as well as why a certain prediction is made.

Predictive models are a much-welcomed tool in the management of MD, but they do indeed need to be interpretable and explainable if they are to find acceptance by the healthcare professionals. For instance, if the model predicts that the patient is at high risk, it would be helpful for the provider to understand which biomarkers—Thai is, CPK levels, or muscle activity from EMG—are contributing most to this risk. Models in Random Forest, through the feature importance metrics, provide this level of transparency.

While this indeed is a research study to provide real-time surveillance for effective intervention, it also has to be established that the model proves not just accurate in predicting it, but also effective in handling continuous streams of data. Architecturally speaking, models like LSTM are really conceived to work on continuous, sequential sets of data, hence in effect providing immediate insights as new streams of data start to accumulate.

This becomes all the more critical for anticipatory surveillance of muscle disease, in which early signs, such as sudden elevations in muscle enzyme levels or decreases in oxygen saturation values, provide immediate guidance for healthcare decisions.

In the case of MD, by the time the scheduled check-ups catch the problem, the damage is done and often irreversible. Real-time predictions mean the system can instantly alert the patients and their health care providers in cases where a patient is at risk, thus enabling timely interventions. LSTM is ideal for such tasks, as it can process continuous new data and predict future trends.

The model is scalable, designed to handle multiple patients in the system. Indeed, Random Forest and SVM models are inherently scalable for handling colossal volumes of data without degradation in performance. It will be quite easy to accommodate their trends for personalized care as more data comes through from the patients. These can then be further fine-tuned by incorporating new data that may reflect evolving understanding or changes in the pathology process. This will not only increase the precision of the predictions but also allow the care strategies of these patients to constantly be optimized.

**Model Training and Validation**

The development of a predictive framework capable of yielding accurate predictions and effective generalization to unknown datasets requires training and validation of the models. This approach begins with the dividing of data into separate subsets for training, validation, and testing, hence the model will learn using one subset, be refined using another, and finally be tested using the last subset.

Patterns within the data are learned in training, and the adjustments made are model-class specific-ArIMA, LSTM, Random Forest, or SVM.

Model learning settings hyperparameter tuning prevents overfitting as well as underfitting; cross-validation: training on a different subset of data and averaging the result along with checking for robustness; evaluation: test model performance on some unseen data sets based on metrics like accuracy, precision, recall, or MAE/RMSE for time-series forecasting.

Techniques like early stopping or pruning reduce the possibility of overfitting, while the increasing complexity of a model addresses underfitting. The aim is to build a model that will give accurate predictions on training data as well as in real-world applications, thus enabling reliable predictions to assist in the monitoring of disease progression and timely interventions in Muscular Dystrophy.

**Real-Time Prediction and Risk Scoring**

Real-time prediction, of course, requires an integral inclusion of new data obtained from wearable technology: CPK levels, EMG signals, $SpO_2$, and heart rate. All the sensors attached systematically collect information at set intervals, and with the availability of new data, the model analyzes this to predict the health condition of the patient. The greatest advantage of real-time forecasting is that it allows healthcare practitioners to be proactive rather than waiting until a complication arises. If the model detects concerning trends-for example, a spike in CPK levels or a drop in oxygen saturation ($SpO_2$)-it could immediately start sending alerts to the patient and the provider. This would allow either an adjustment in the medication regimen or behavioral advice to prevent these health threats from worsening in time. In particular, for MD management, biomarkers such as CPK levels receive special consideration, as they are a direct measure of muscle damage. Thus, if CPK levels exceed a certain threshold (for example, 1000 U/L), the risk score increases, indicating a greater likelihood of severe muscle damage. On the other hand, EMG signals offer real-time information about muscle activity and contribute to the risk score by potentially showing functional decline.

Heart rate and $SpO_2$ are also integral components of the scoring system, where diminished oxygen saturation coupled with heightened heart rate variability indicates an elevated risk for cardiovascular complications.

The risk score exhibits dynamism, as it adapts in response to newly acquired data, thereby offering a contemporary evaluation of the patient's health status. This characteristic enables the healthcare team to modify treatment strategies in accordance with deteriorating trends. Should the model identify an increase in CPK levels or other relevant biomarkers, the risk score may be revised, instigating more regular monitoring, alterations in medication, or a recommendation for further medical attention. The risk scoring framework further enables personalized recommendations for patients. The recommendations are thus personalized for each patient based on the current state of their health condition, as expressed through their risk score and the predictive insights returned by the model. In addition, the alerts are further communicated to patients and healthcare professionals. Individuals receiving healthcare could obtain push notifications through a mobile application, whereas healthcare professionals might receive alerts via an online dashboard that concurrently monitors other essential metrics. The adaptive system, in its process for real-time prediction and risk scoring, involves continuous updates of predictions and risk scores with newly acquired data from wearables. Therefore, the continuous acquisition of data ensures that all the time the system operates using the latest information, creating accurate evaluations at all moments. Real-time forecasting and risk assessment constitute integral components in the effective management of Muscular Dystrophy (MD). By continuously analyzing data from wearable sensors and incorporating it into an adaptive risk scoring system, healthcare practitioners are provided with immediate insights into a patient's health condition, enabling prophylactic interventions before complications arise. These predictive models not only forecast disease progression but also offer personalized care recommendations, helping to tailor treatment plans to each patient's unique needs. This proactive approach is essential for improving patient outcomes and ensuring that care is both timely and effective.

## V. Conclusion and Future Work

This research showed the drastic effect of predictive modeling in the management of muscular dystrophy by using real-time data collection, advanced machine learning techniques, and pre-emptive intervention strategies. The article details a framework for continuous surveillance by integrating wearable Internet of Things (IoT) devices, cloud-based analytics tools, and predictive models like ARIMA, LSTM, Random Forest, and Support Vector Machine (SVM), hence allowing tailored care that would significantly improve clinical outcomes and the quality of life for the patients.

Similarly, using time-series forecasting through ARIMA allows MD progression, for example, changes in CPK levels, to be predicted, and interventional strategies can be provided for the same. LSTM networks learn the long-term patterns of sequential data, including EMG signals and heart rate, which prove valuable in the insight of disease progression. At the same time, Random Forest and SVM will classify patients according to risk, enabling more targeted and timely treatment interventions based on dynamic risk assessments.

The proposed system includes the real-time forecasting and adaptive risk assessment mechanisms developed in this study. These mechanisms continuously evaluate new input data, updating risk scores and giving personalized care recommendations so that health professionals may intervene early. More than that, the system provides opportunities for proactive monitoring by healthcare providers and, on the other hand, allows patients to be more engaged in their health management with immediate insight into and alerts about their conditions.

While the system introduced in this study has significant potential to bring about better management of MD, numerous further research directions exist. One key area is integrating more data sources into the model, including genetic data and lifestyle variables, to increase the accuracy of the model's prediction and allow for more personalized strategies of care. A study of more advanced machine learning methods, such as deep learning and reinforcement learning, could potentially improve adaptability and predictive stability over time in a system.

Such real-time monitoring systems, as developed in this research, are going to be increasingly necessary for the management of chronic diseases. Coming studies will show how these latest IoT advancements, including cloud computing and AI models, could further optimize predictive modeling to offer even more effective treatment protocols tailored for each MD patient. These innovations will not only help improve MD care but might also lay the foundation for the wider application of predictive medicine in other chronic diseases, thereby enlarging the potential impact of such an approach within healthcare.